\documentclass{article}
\usepackage{spconf,amsmath,epsfig}

\let\OLDthebibliography\thebibliography
\renewcommand\thebibliography[1]{
  \OLDthebibliography{#1}
  \setlength{\parskip}{0pt}
  \setlength{\itemsep}{0pt plus 0.3ex}
}

\usepackage{algorithm}
\usepackage{algorithmic}
\usepackage{epsfig}
\usepackage{amsmath}
\usepackage{amssymb}
\usepackage{subfigure}
\usepackage{makecell}
\usepackage{comment}
\usepackage{romannum}
\usepackage{mathtools}
\usepackage{commath}
\usepackage{multirow}
\usepackage{graphicx}
\usepackage[table,xcdraw]{xcolor}

\begin{document}\sloppy

\def\x{{\mathbf x}}
\def\L{{\cal L}}

\title{Learning Facial Liveness Representation \\ for Domain Generalized Face Anti-Spoofing}
%
\name{
\parbox{\linewidth}{\centering
      Zih-Ching Chen$^{\star \dag}$,
      Lin-Hsi Tsao$^{\star \dag}$,
      Chin-Lun Fu$^{\star \dag}$,\thanks{$^{\star}$ - indicates equal contribution.}
      Shang-Fu Chen$^{\dag \ddag}$,
      Yu-Chiang Frank Wang$^{\dag}$%
    }%
}
\address{$^{\dag}$ Graduate Institute of Communication Engineering, National Taiwan University, Taiwan\\
$^{\ddag}$ Telecommunication Laboratories, Chunghwa Telecom Co., Ltd., Taiwan}
\maketitle

\begin{abstract}
Face anti-spoofing (FAS) aims at distinguishing face spoof attacks from the authentic ones, which is typically approached by learning proper models for performing the associated classification task. In practice, one would expect such models to be generalized to FAS in different image domains. Moreover, it is not practical to assume that the type of spoof attacks would be known in advance. In this paper, we propose a deep learning model for addressing the aforementioned domain-generalized face anti-spoofing task. In particular, our proposed network is able to disentangle facial liveness representation from the irrelevant ones (i.e., facial content and image domain features). The resulting liveness representation exhibits sufficient domain invariant properties, and thus it can be applied for performing domain-generalized FAS. In our experiments, we conduct experiments on five benchmark datasets with various settings, and we verify that our model performs favorably against state-of-the-art approaches in identifying novel types of spoof attacks in unseen image domains.
\end{abstract}


\section{Introduction}

Face recognition technology has been widely applied in many interactive intelligent systems such as automated teller machines (ATMs), mobile payments, and entrance guard systems, due to their convenience and remarkable accuracy. However, face recognition systems are still vulnerable to presentation attacks ranging from print attacks, video replay attacks, and 3D facial mask attacks, etc. Therefore, face anti-spoofing (FAS) plays a crucial role in securing the robustness of face recognition systems.

Over the past few years, different FAS methods have been proposed by researchers. Assuming inherent disparities between live and spoof faces, studies have handled this problem from the perspective of detecting texture in color space~\cite{boulkenafet2015face}, image distortion~\cite{wen2015face}, temporal variation~\cite{shao2017deep}, or deep semantic features~\cite{yang2014learn, patel2016cross}. Although promising results have been obtained by these methods, to perform FAS on unseen image domains which are not observed during training remains a challenging task. The performance of these FAS methods trained from a particular source domain would drop dramatically when different backgrounds, subjects, or shooting devices are encountered in a different target domain of interest.

To address the domain shift problem mentioned above, studies have exploited auxiliary information, such as face depth~\cite{liu2018learning}, for distinguishing live and spoof faces. However, these approaches still have their limitations since they highly depend on the accuracy of estimated auxiliary information. Therefore, researchers start to improve the robustness of FAS from the perspective of domain generalization, which aims to learn a generalized feature space by aligning the distributions among multiple source domains. Shao~\cite{shao2019multi} proposed a multi-adversarial deep domain generalization (MADDG) framework to derive domain-invariant feature spaces for real and fake images with a dual-force triplet mining constraint. Extended from MADDG, Jia~\cite{jia2020single} proposed a single-side domain generalization (SSDG) learning framework that groups spoof types across domains together with a triplet-mining algorithm for the purpose of domain generalization. However, the generalization ability of the mentioned approaches might be limited. This is because the existing methods typically do not distinguish between domain-independent facial liveness representations and the domain-dependent ones which are irrelevant to FAS~\cite{wang2020cross}. 

In light of the above issue, researchers handled FAS from the aspect of feature disentanglement. A multi-domain disentangled representation learning method is proposed by~\cite{wang2020cross}, aiming to obtain more discriminative liveness features by disentangling domain-invariant representations from an image. While~\cite{wang2020cross} have disentangled the domain-independent FAS cues from the domain-dependent representations, their approach might not be able to generalize to real-world scenarios, in which novel (i.e., unseen) spoof attacks might be presented. 
Although existing works focus on disentangling domain-relevant features from other features, their extracted liveness features still contain facial content information, which is irrelevant to liveness information. Thus, the generalization ability of their methods to handle unseen spoof attacks is still limited.

In this paper, we address the domain-generalized FAS problem by learning domain-invariant facial liveness representation. We not only aim to handle FAS in unseen data domains but unseen spoof attacks can also be detected, which makes our proposed model more practical. As detailed later, we present a representation disentanglement framework, which is designed to extract facial liveness, content, and image domain representations. More specifically, the liveness representation describes information for liveness detection. On the other hand, the latter two representations, i.e., the facial content and image domain representations are viewed as liveness-invariant features. The disentanglement of such features from the liveness features allows our model to better perform FAS in unseen domains with novel spoof attacks.
The contributions of this work can be highlighted below:
\begin{itemize}
  \item  We propose a representation disentanglement network for domain-generalized face liveness detection, which is able to recognize novel spoof attacks in unseen domains/datasets during inference.
  \item  Our proposed network is designed to extract facial liveness, content, and image domain representations. While the liveness representation would be utilized for FAS, the latter two are liveness-invariant.
  \item  We conduct experiments on multiple FAS datasets in various settings, and confirm that our method performs favorably against state-of-the-art approaches in detecting novel spoof attacks in unseen image domains.
\end{itemize}

\begin{figure*}[t]
\begin{center}
\includegraphics[width=0.7\textwidth]{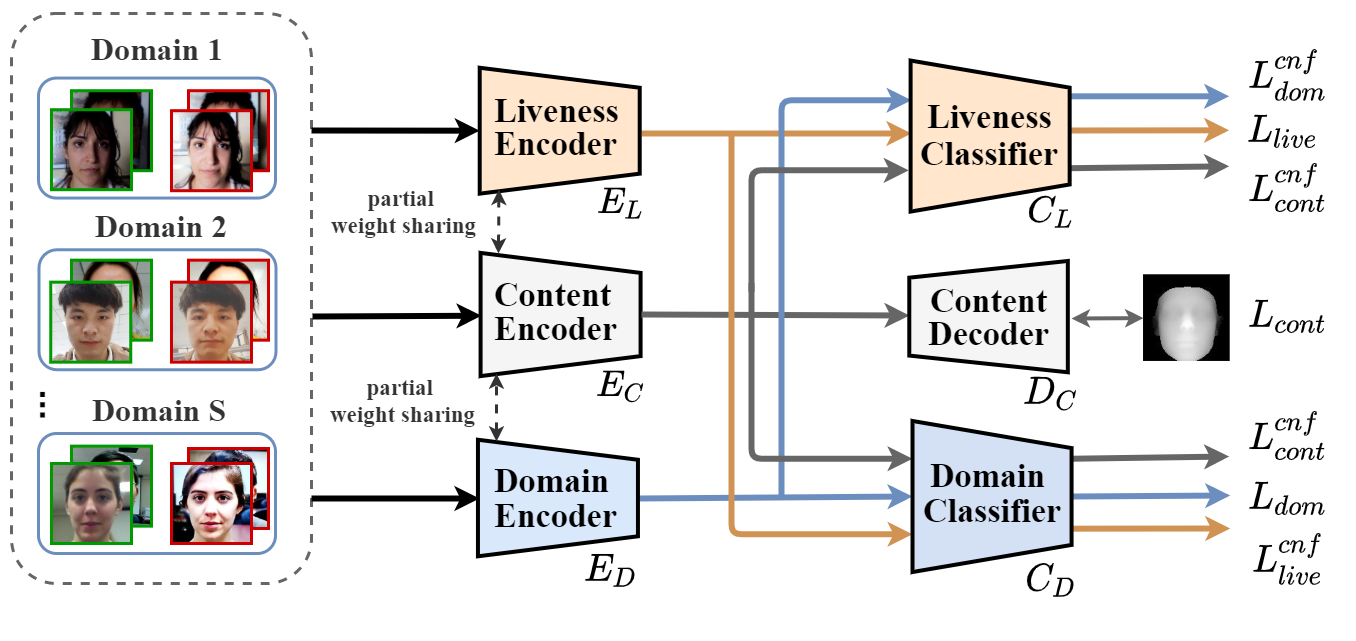}
\vspace{-0.6cm}
\end{center}
   \caption{Overview of our proposed network architecture. Our network aims to extract liveness, facial content, and image domain representations from facial input images. This is realized by the learning of liveness encoder $E_L$, content encoder $E_C$, and domain encoder $E_D$, respectively. To further ensure the disentanglement of liveness-irrelevant information, liveness classifier $C_L$, content decoder $D_C$, and domain classifier $C_D$ are jointly deployed. Once the training is complete, one can apply $E_L$ and $C_L$ for domain-generalized FAS.}
\label{fig:archi}
\end{figure*}
\section{Proposed Method}
\subsection{Problem Definition and Annotations}
For the sake of completeness, we first define the setting and notations considered in this paper. During training, we have face images in \emph{S} different source domains denoted as $X = \{X_1, X_2, ..., X_S \} $ and the corresponding binary real/fake labels denoted as $Y = \{Y_1, Y_2, ..., Y_S \} $. For the $i${th} source domain, we have ${N_i}$ images, i.e., $X_i = \{x_{i,j}\}_{j=1}^{N_i}$, and the associated labels $Y_i = \{y_{i,j}\}_{j=1}^{N_i}$. As for the inference stage, liveness of facial images in a disjoint and unseen target domain (with seen attacks and unseen attacks) will be determined accordingly.

As shown in Fig.~\ref{fig:archi}, our proposed network architecture has three encoders for handling the facial input images: liveness encoder $E_{L}$, content encoder $E_{C}$, and domain encoder $E_{D}$. The liveness encoder $E_{L}$ is designed to extract liveness representation, followed by a liveness classifier $C_{L}$ for performing FAS prediction. The content encoder $E_{C}$ extracts the facial content representation from the input, while the subsequent decoder $D_{C}$ is deployed for reconstruction guarantees. As for the encoder $E_{D}$, it extracts the domain representations from the training input images, so that the domain classifier $C_{D}$ would classify the image domain accordingly. Once the joint learning of the above network modules is complete, one can simply take the liveness encoder $E_{L}$ and the liveness classifier $C_{L}$ for inference.

\subsection{Learning Liveness-Irrelevant Representation}
In order to address domain-generalized FAS, we propose to extract facial content and image domain features from the liveness representation. Since the disentangled content and image domain features are not utilized for FAS, they can be viewed as liveness-irrelevant representations of face images.

As depicted in Fig.~\ref{fig:archi}, the above two types of features are extracted by the content encoder $E_{C}$ and domain encoder $E_{D}$. For the content encoder, it is expected to retrieve facial content information, which is not regarding the authenticity of the face input image. In our work, we specifically consider content information described by PRNet~\cite{feng2018joint}, which is known for face alignment information by observing a face image. The idea is that spoof image or not, the facial input image is expected to contain facial contour and landmark information, which suggest the recovery of the corresponding face alignment. Thus, given the $j$th training image from source domain $i$, the encoded content feature of $E_{C}(x_{i,j})$ would serve as the input to the content decoder $D_{C}$, which is designed to recover the feature produced by a pre-trained PRNet $\Phi(\cdot)$. In other words, we calculate the following \textit{content loss} $L_{cont}$ for updating both $E_{C}$ and $D_{C}$:

\begin{equation}
\normalsize L_{cont} = \sum_{i=1}^{S} \sum_{j=1}^{N}\|D_{C}(E_{C}(x_{i,j}))-\Phi(x_{i,j})\|_2^2\label{eq:1}.
\end{equation}

While the calculation of~\eqref{eq:1} ensures $E_{C}$ and $D_{C}$ for extracting and recovering facial content information, we need additional supervision to ensure the derived content features $E_{C}(x_{i,j})$ does not contain either liveness or image domain information. As a results, with the deployment of the liveness classifier $C_{L}$ and domain classifier $C_D$, we choose to calculate the following \textit{content confusion loss} $L^{cnf}_{cont}$: 

\begin{equation}
\begin{aligned}
\normalsize {L^{cnf}_{cont}}=\sum_{i=1}^{S} \sum_{j=1}^{N}(&\|C_{L}(E_{C}(x_{i,j}))-{\frac{1}{2}}\|_2^2 \\
+&\|C_{D}(E_{C}(x_{i,j}))-{\frac{1}{S}}\|_2^2).\label{eq:2}
\end{aligned}
\end{equation}

\noindent The first and second terms in~\eqref{eq:2} are to confuse the liveness and domain classifiers, respectively. It is worth noting that, $C_{L}$ predicts the binary liveness label, and $C_{D}$ outputs the domain label (out of $S$). With the observation of this content confusion loss, the training of our proposed framework would further ensure the content encoder $E_{C}$ to produce liveness and domain-irrelevant features.

As for the learning of image domain features, the modules of domain encoder $E_C$ and domain classifier $C_D$ are deployed for achieving this goal. Following the above example, assume that we have the $j$th training image from source domain $i$, the encoded domain feature $E_D(x_{i,j})$ is expected to describe illumination, image quality, etc., information. Thus, the subsequent domain classifier $C_{D}$ is designed to recognize the image domain $i$ by observing such domain features. In our work, we calculate the \textit{domain loss} $L_{dom}$ for updating both $E_{D}$ and $C_{D}$ as follows:

\begin{equation}
\normalsize L_{dom}=-\sum_{i=1}^{S} \sum_{j=1}^{N}m_{i} *log(C_{D}(E_D(x_{i,j}))).\label{eq:3}
\end{equation}

\noindent In the above equation, $m_i$ denotes the ground truth one-hot vector representing the domain label. 

Similar to the design of liveness-irrelevant facial content features, we now discuss how we further ensure that the learning of $E_{D}$ and $C_{D}$ would not contain liveness information. With the deployment of the liveness classifier $C_L$ in Fig.~\ref{fig:archi}, we have $C_L$ take the encoded domain features $E_D(x_{i, j})$. To enforce the disentanglement of liveness-relevant information, the liveness classifier $C_L$ is not expected to perform FAS on $E_D(x_{i, j})$. Therefore, we calculate the \textit{domain confusion loss} $L^{cnf}_{dom}$, which is defined below:

\begin{equation}
\normalsize L^{cnf}_{dom}=\sum_{i=1}^{S}\sum_{j=1}^{N}\|C_{L}(E_D(x_{i,j}))-{\frac{1}{2}})\|_2^2.\label{eq:4}
\end{equation}

With facial content and domain losses, together with the corresponding confusion losses, our proposed framework allows us to disentangle liveness-irrelevant features from the input images. The deployment of $E_C$, $E_D$, $D_C$, and $C_D$ would also facilitate the learning of liveness representation, as we discuss next. 

\subsection{Learning Domain-Invariant Liveness Representation}

To address domain-generalized FAS, learning of domain-invariant liveness representation from the input images would be the major component of our proposed framework. With facial content and image domain features properly disentangled from the input image, we now discuss how the extraction of liveness representation is realized by our network so that the derived features can be applied to detect novel spoof attacks in unseen target domains.

As depicted in Fig.~\ref{fig:archi}, we have liveness encoder $E_L$ and the associated classifier $C_L$ deployed in our framework. Given the $j$th training image from source domain $i$, we have the encoded liveness feature $E_L(x_{i,j})$ expected to describe the liveness (i.e. real/fake) information. The subsequent liveness classifier $C_L$ is designed to determine whether the input face image is real or fake depending on the feature $E_L(x_{i,j})$. To better separate real and fake facial images, we adopt the simplified Large Margin Cosine Loss (LMCL) function~\cite{wang2018cosface} as the objective, which calculates intra/inter-class angular distances for the corresponding input images with a predetermined margin $m$. For the sake of clarity, we disregard the domain index $i$ for the input image; thus, \textit{liveness loss} $L_{live}$ for updating $E_L$ and $C_L$ is calculated as follows:   

\begin{table*}[ht]
\vspace*{-5pt}
\begin{center}
\resizebox{\textwidth}{26mm}{
\scalebox{1.0}{
\begin{tabular}{c|c|c|c|c|c|c|cc}
\hline
                                  & \multicolumn{2}{c|}{{\color[HTML]{333333} \textbf{O\&C\&I to M}}} & \multicolumn{2}{c|}{\textbf{O\&M\&I to C}} & \multicolumn{2}{c|}{\textbf{O\&C\&M to I}} & \multicolumn{2}{c}{\textbf{I\&C\&M to O}}            \\ \cline{2-9} 
\multirow{-2}{*}{\textbf{Method}} & HTER(\%)                        & AUC(\%)                         & HTER(\%)            & AUC(\%)              & HTER(\%)             & AUC(\%)             & \multicolumn{1}{c|}{HTER(\%)}       & AUC(\%)        \\ \hline \hline

Auxiliary(Depth Only)             & 29.14                           & 71.69                           & 22.72               & 85.88                & 33.52                & 73.15               & \multicolumn{1}{c|}{30.17}          & 77.61          \\
Auxiliary(All)~\cite{liu2018learning}                    & 27.60                           & -                               & -                   & -                    & 28.40                & -                   & \multicolumn{1}{c|}{-}              & -              \\
MMD-AAE~\cite{li2018domain}                           & 31.58                           & 75.18                           & 27.08               & 83.19                & 44.95                & 58.29               & \multicolumn{1}{c|}{40.98}          & 63.08          \\
MADDG~\cite{shao2019multi}                             & 17.69                           & 88.06                           & 24.50               & 84.51                & 22.19                & 84.99               & \multicolumn{1}{c|}{27.98}          & 80.02          \\
RFM~\cite{shao2020regularized}                               & 13.89                           & 93.98                           & 20.27               & 88.16                & 17.30                & 90.48               & \multicolumn{1}{c|}{16.45}          & 91.16          \\
SSDG-M~\cite{jia2020single}                            & 16.67                           & 90.47                           & 23.11               & 85.45                & 18.21                & 94.61               & \multicolumn{1}{c|}{25.17}          & 81.83          \\
SSDG-R~\cite{jia2020single}                            & \textbf{7.38}                   & 97.17                           & 10.44               & 95.94                & 11.71                & \textbf{96.59}      & \multicolumn{1}{c|}{\textbf{15.61}} & 91.54          \\
Cross~\cite{wang2020cross}                             & 17.02                           & 90.10                           & 19.68               & 87.43                & 20.87                & 86.72               & \multicolumn{1}{c|}{25.02}          & 81.47          \\
DRDG~\cite{liu2021dual}                              & 12.43                           & 95.81                           & 19.05               & 88.79                & 15.56                & 91.79               & \multicolumn{1}{c|}{15.63}          & 91.16          
\\ \hline
\hline
\textbf{Ours}                     & 7.50                            & \textbf{97.45}                  & \textbf{9.80}       & \textbf{96.82}       & \textbf{11.38}       & 94.90               & \multicolumn{1}{c|}{16.70}          & \textbf{91.83} \\ \hline

\end{tabular}}}
\end{center}
\vspace{-0.3cm}
\caption{Comparisons of FAS in unseen domains in terms of HTER and AUC. For example, O\&C\&I to M denotes that the model is trained on the datasets of Oulu, CASIA, \&  Idiap, and is evaluated on MSU. Note that the attack types are print and replay in both source and target domains.}
\label{tab:und}
\end{table*}
\begin{equation}
\begin{aligned}
\normalsize &L_{live}= -\sum _{j=1}^{N}log(\frac{e^{\alpha(W^T_{y_j}E_L(x_{j})-m)}}{e^{\alpha(W^T_{y_j}E_L(x_j)-m)} + e^{\alpha(W^T_{1-y_j}E_L(x_{j}))}}),\label{eq:5}
\end{aligned}
\end{equation} 
where $\alpha$ is a hyperparameter, and $W=\{W_0,W_1\}$ denotes the parameters of liveness classifier $C_L$. We note that, $W_0$ represents the model parameter for label $y_j = 0$ (i.e., spoof attack), while $W_1$ denotes that for $y_j = 1$ (i.e., real face). And, the margin $m$ is introduced in Eq.~\eqref{eq:5}, so that the separation between liveness representations derived from real and fake images can be further enforced.

To further ensure that the liveness feature $E_L(x_{i,j})$ does not contain any image domain information, we utilize the aforementioned domain classifier $C_D$, and calculate the \textit{liveness confusion loss} $L^{cnf}_{live}$ as follows: 

\begin{equation}
\normalsize {L^{cnf}_{live}}=\sum_{i=1}^S \sum_{j=1}^{N}\|C_{D}(E_L(x_{i,j}))-\frac{1}{S})\|_2^2.\label{eq:6}
\end{equation}

\noindent With both liveness classification and confusion losses, we are able to train our proposed framework for disentangling domain-invariant liveness representation.

Together with Eq.~\eqref{eq:5} and Eq.~\eqref{eq:6}, we maximize $W^T_{y_j}E_L(x_j)$ and minimize $W^T_{1-y_j}E_L(x_j)$, which encourages the separation between real and fake images (with margin $m$) across domains. On the other hand, maximization of $W^T_{y_j}E_L(x_j)$ further implies the suppression of intra-class variation for images of the same label. In other words, the learned $W_1$ and $W_0$ would represent the \textit{prototypes} of domain-invariant liveness representations of real and fake images, respectively. With the disentanglement of image domain information, our learning of liveness representation would not only separate real face images and spoof attacks but also enforce the minimization of the associated intra-class variations. Therefore, the generalization of our model to novel spoof attacks across different domains can be expected.

The overall objectives, together with the detailed training process, are summarized in the Algorithm 1 of our supplementary materials. Once the training of our network architecture is complete, only $E_L$ and $C_L$ are utilized for performing domain generalized FAS. That is, the liveness encoder $E_L$ is applied to extract the domain-invariant liveness representation, which is fed into $C_L$ for FAS prediction.


\section{Experiment}
\subsection{Experimental Settings}
 Five public face anti-spoofing datasets are utilized to evaluate the effectiveness of our method: OULU-NPU~\cite{boulkenafet2017oulu} (denoted as O), CASIA-FASD~\cite{zhang2012face} (denoted as C), Idiap Replay-Attack~\cite{chingovska2012effectiveness} (denoted as I), MSU-MFSD~\cite{wen2015face} (denoted as M), and CelebA-Spoof~\cite{zhang2020celeba} (denoted as Cb). 

\begin{table*}[ht]
\vspace*{-5pt}
\centering
\scalebox{1.0}{
\begin{tabular}{c|c|c|c|c|c|c|cc}
\hline
                                  & \multicolumn{2}{c|}{{\color[HTML]{333333} \textbf{O\&C\&I to Cb}}} & \multicolumn{2}{c|}{\textbf{O\&M\&I to Cb}} & \multicolumn{2}{c|}{\textbf{O\&C\&M to Cb}} & \multicolumn{2}{c}{\textbf{I\&C\&M to Cb}}           \\ \cline{2-9} 
\multirow{-2}{*}{\textbf{Method}} & HTER(\%)                         & AUC(\%)                         & HTER(\%)             & AUC(\%)              & HTER(\%)             & AUC(\%)              & \multicolumn{1}{c|}{HTER(\%)}       & AUC(\%)        \\ \hline\hline
MADDG~\cite{shao2019multi}                             & 42.46                            & 62.26                           & 44.96                & 57.01                & 50.08                & 57.18                & \multicolumn{1}{c|}{48.92}          & 51.86          \\
RFM~\cite{shao2020regularized}                               & 41.49                            & 60.19                           & 43.32                & 65.96                & 42.83                & 59.37                & \multicolumn{1}{c|}{35.93}          & 67.32          \\
SSDG-M~\cite{jia2020single}                            & 41.19                            & 60.91                           & 35.25                & 65.96                & 36.40                & 66.66                & \multicolumn{1}{c|}{35.02}          & 69.19          \\
SSDG-R~\cite{jia2020single}                            & 20.29                     & 86.87                           & \textbf{20.58}       & 86.54                & 25.05                & 82.11                & \multicolumn{1}{c|}{19.86}          & 88.58          \\ \hline\hline
\textbf{Ours}                     & \textbf{19.42}                       & \textbf{88.17}                 & 20.60                & \textbf{86.93}       & \textbf{22.32}       & \textbf{85.49}       & \multicolumn{1}{c|}{\textbf{16.22}} & \textbf{90.85} \\ \hline
\end{tabular}}
\caption{Comparisons of FAS in unseen domains with novel spoof attacks in terms of HTER and AUC. Note that the attack types are print and replay in the source domains, while Cb contains the spoof attack of 3D masks for testing.}
\label{tab:att}
\end{table*}
For the architecture of our model, we have ResNet-18~\cite{he2016deep} pre-trained on ImageNet for the encoders $E_L$, $E_C$ and $E_D$, where the weights of the first layer are shared. The statistics of each dataset and the details of the architecture are shown in Table A. and Table B. of the supplementary materials, respectively.
For the evaluation metrics, we have the Half Total Error Rate (HTER) and Area under the Curve of ROC (AUC) following~\cite{shao2019multi, jia2020single, wang2020cross, liu2021dual} in all experiments. 


\subsection{FAS in Unseen Target Domains}
Following~\cite{shao2019multi, jia2020single, wang2020cross, liu2021dual}, we utilize four public datasets, i.e., O, C, M, and I, to evaluate the effectiveness of our model adapting to unseen datasets. We select three datasets as the source domains and the remaining one as the target domain.

As shown in Table~\ref{tab:und}, our approach achieved impressive results and performed against all the existing FAS approaches. This demonstrates that the liveness encoder $E_L$ in our proposed model is able to extract generalized features for the unseen domains by disentangling the liveness-irrelevant information. While recent approaches including MADDG and SSDG consider extracting domain-generalized liveness features, the generalization ability of their methods is still limited. This is because that their extracted liveness features still contain facial information, which is irrelevant to liveness information.

\subsection{FAS with Novel Spoof Attacks in Unseen Target Domains} 
\begin{table*}[ht]
\vspace*{-8pt}
\centering
\begin{tabular}{c|c|c|c|c|c|c|cc}
\hline
                                  & \multicolumn{2}{c|}{{\color[HTML]{333333} \textbf{O\&C\&I to Cb}}} & \multicolumn{2}{c|}{\textbf{O\&M\&I to Cb}} & \multicolumn{2}{c|}{\textbf{O\&C\&M to Cb}} & \multicolumn{2}{c}{\textbf{I\&C\&M to Cb}}            \\ \cline{2-9} 
\multirow{-2}{*}{\textbf{Method}} & HTER(\%)                        & AUC(\%)                         & HTER(\%)            & AUC(\%)              & HTER(\%)             & AUC(\%)             & \multicolumn{1}{c|}{HTER(\%)}       & AUC(\%)        \\ \hline\hline
Baseline                    & 34.41                           & 72.81                           & 45.01               & 57.53                & 33.36                & 72.9               & \multicolumn{1}{c|}{38.30}          & 66.74         \\
Baseline + $E_C$, $D_C$                  & 33.26                           & 73.23                           & 41.94               &61.63                & 28.13                &77.35               & \multicolumn{1}{c|}{39.15}          & 68.78          \\ 
Baseline + $E_D$, $C_D$                   & 27.66                           & 79.40                           & 34.64               & 69.59                & 25.99                & 80.62               & \multicolumn{1}{c|}{26.15}          & 82.54          \\ \hline\hline
\textbf{Ours}                     & \textbf{19.42}                   & \textbf{88.17}                  & \textbf{20.60}       & \textbf{86.93}       & \textbf{22.32}       & \textbf{85.49}      & \multicolumn{1}{c|}{\textbf{16.22}} & \textbf{90.85} \\ \hline
\end{tabular}
\caption{Analysis of our network architecture design. Note that Baseline denotes the learning of only the liveness encoder and liveness classifier. The modules for disentangling liveness-irrelevant representations (i.e., $E_C$\&$D_C$ for content and $E_D$\&$C_D$ for image domain) are assessed, which are shown to support the effectiveness of our full model.}
\label{tab:abl}
\end{table*}
We evaluate the effectiveness of our model adapting to a real-world scenario where the model encounters unseen spoof attacks which are not observed during training. Five datasets are used in this setting: O, C, I, M, and Cb. We choose Cb as the target domain since Cb contains a unique attack type, i.e. 3D mask, which does not appear in other datasets. 

As shown in Table~\ref{tab:att}, we can observe that our approach outperforms existing FAS approaches in all of the four tasks. Our proposed method can be better generalized to unseen spoof attacks because it learns the domain-invariant liveness representations, which are able to generalize to unseen target domains. 
As explained in the \textit{liveness loss} $L_{live}$, our liveness classifier $C_L$ could learn a better liveness feature prototype after seeing different spoof attacks from the training data. Therefore, the learned fake prototype of the liveness classifier can generalize better to unseen spoof attacks.

\subsection{Qualitative Analysis}
As discussed in the previous sections, recent DG approaches, including SSDG, focus on extracting domain-invariant features but neglect that the derived liveness features may still contain irrelevant facial content information. On the other hand, our method disentangles such facial content for better generalization performances.
To verify the effectiveness of our disentanglement model, we utilize the Grad-CAM~\cite{selvaraju2017grad} algorithm to obtain the class activation map visualizations, where the activation map indicates the regions that the model attends on when performing domain-generalized FAS. 
 
\begin{figure}[t]
    \begin{center}
    \includegraphics[width=\linewidth]{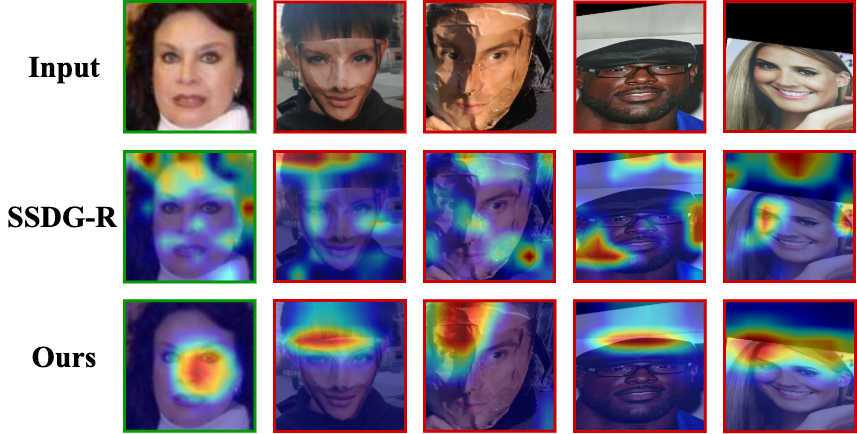}
    \vspace{-0.5cm}
    \caption
    {FAS visualization examples on real and spoof attacks (in green and red bounding boxes, respectively) using O\&M\&I to Cb. Note that the Grad-CAM is utilized to visualize the activation maps for real and fake images. Comparing to SSDG-R, our method offers explainable attention maps.
    \vspace{-0.3cm}}
    \end{center}
    \label{fig:vis}
\end{figure}

We compare our model to SSDG-R and show the visualization results in Figure~\ref{fig:vis}. The first column of the figure shows the visualization result of a real face and the other columns present the results of spoof faces. In the first column, SSDG-R focused on the background of the face image while our proposed model concentrated on the liveness part of the face image. For the spoof image in the third column, with the facial content features disentangled, our proposed model concentrated on the sunken part of the 3D mask, which could be regarded as important liveness information. On the other hand, SSDG-R focuses more on the liveness-irrelevant facial contours and background. The visualization results support that disentanglement of the facial content enforce our model on the liveness-related part of the face images and therefore facilitate the domain-generalized FAS.

\subsection{Ablation Study}
As listed in Table~\ref{tab:abl}, the baseline model (in the first row) contains only the liveness encoder $E_L$ and liveness classifier $C_L$. The second and the third row shows the results when the facial content disentanglement and domain disentanglement is applied, respectively. The last row shows the results of our proposed model, i.e., both content and domain disentanglement are applied.

We can see from the first two rows that the model with the disentanglement of facial content performs better than the baseline. This confirms that such disentanglement helps our model to focus on the liveness cues in the images instead of liveness-irrelevant facial content. Comparing the first and third row, the domain disentanglement mechanism helps our model extract domain-invariant liveness features that can be applied on the unseen target domain and therefore brings significant improvements. With both facial content and domain feature disentanglement are utilized, our proposed method achieved the best performance in all tasks.

\section{Conclusion}
In this paper, we address the challenging task of domain-generalized FAS problem, in which novel spoof attacks in unseen target domains need to be identified. Based on the idea of representation disentanglement, we present a network architecture that is able to extract facial liveness, content, and domain features. Aiming at performing domain-generalized FAS, the facial liveness representation exhibits domain invariant properties while the content and domain representations are viewed as liveness-irrelevant features, whose derivations are enforced by our network module and objective designs. Extensive experiments on benchmark datasets demonstrated the effectiveness of our proposed network, which shows promising domain generalization in addressing FAS, including the ability to handle novel types of spoof attacks during inference. Since the current proposed model only observes images as inputs, taking video inputs and thus dealing with visual-temporal information would be among our future research directions for domain-generalized FAS.\\

\noindent\textbf{Acknowledgement}
This work is supported in part by the Telecommunication Laboratories of Chunghwa Telecom TL-110-L402. We also thank National Center for High-performance Computing (NCHC) for providing computational and storage resources.
%
\bibliographystyle{IEEEbib}
\bibliography{icme}

\begin{thebibliography}{10}

\bibitem{boulkenafet2015face}
Zinelabidine Boulkenafet, Jukka Komulainen, and Abdenour Hadid,
\newblock ``Face anti-spoofing based on color texture analysis,''
\newblock in {\em 2015 ICIP}. IEEE, 2015.

\bibitem{wen2015face}
Di~Wen, Hu~Han, and Anil~K Jain,
\newblock ``Face spoof detection with image distortion analysis,''
\newblock {\em IEEE Transactions on Information Forensics and Security}, vol.
  10, no. 4, 2015.

\bibitem{shao2017deep}
Rui Shao, Xiangyuan Lan, and Pong~C Yuen,
\newblock ``Deep convolutional dynamic texture learning with adaptive
  channel-discriminability for 3d mask face anti-spoofing,''
\newblock in {\em 2017 IJCB}. IEEE, 2017, pp. 748--755.

\bibitem{yang2014learn}
Jianwei Yang, Zhen Lei, and Stan~Z Li,
\newblock ``Learn convolutional neural network for face anti-spoofing,''
\newblock {\em arXiv preprint arXiv:1408.5601}, 2014.

\bibitem{patel2016cross}
Keyurkumar Patel, Hu~Han, and Anil~K Jain,
\newblock ``Cross-database face antispoofing with robust feature
  representation,''
\newblock in {\em Chinese Conference on Biometric Recognition}. Springer, 2016,
  pp. 611--619.

\bibitem{liu2018learning}
Yaojie Liu, Amin Jourabloo, and Xiaoming Liu,
\newblock ``Learning deep models for face anti-spoofing: Binary or auxiliary
  supervision,''
\newblock in {\em CVPR}, 2018, pp. 389--398.

\bibitem{shao2019multi}
Rui Shao et~al.,
\newblock ``Multi-adversarial discriminative deep domain generalization for
  face presentation attack detection,''
\newblock in {\em CVPR}, 2019, pp. 10023--10031.

\bibitem{jia2020single}
Yunpei Jia et~al.,
\newblock ``Single-side domain generalization for face anti-spoofing,''
\newblock in {\em CVPR}, 2020, pp. 8484--8493.

\bibitem{wang2020cross}
Guoqing Wang et~al.,
\newblock ``Cross-domain face presentation attack detection via multi-domain
  disentangled representation learning,''
\newblock in {\em CVPR}, 2020, pp. 6678--6687.

\bibitem{feng2018joint}
Yao Feng et~al.,
\newblock ``Joint 3d face reconstruction and dense alignment with position map
  regression network,''
\newblock in {\em ECCV}, 2018, pp. 534--551.

\bibitem{wang2018cosface}
Hao Wang et~al.,
\newblock ``Cosface: Large margin cosine loss for deep face recognition,''
\newblock in {\em CVPR}, 2018.

\bibitem{li2018domain}
Haoliang Li et~al.,
\newblock ``Domain generalization with adversarial feature learning,''
\newblock in {\em CVPR}, 2018, pp. 5400--5409.

\bibitem{shao2020regularized}
Rui Shao, Xiangyuan Lan, and Pong~C Yuen,
\newblock ``Regularized fine-grained meta face anti-spoofing,''
\newblock in {\em AAAI}, 2020.

\bibitem{liu2021dual}
Shubao Liu et~al.,
\newblock ``Dual reweighting domain generalization for face presentation attack
  detection,''
\newblock {\em arXiv preprint arXiv:2106.16128}, 2021.

\bibitem{boulkenafet2017oulu}
Zinelabinde Boulkenafet et~al.,
\newblock ``Oulu-npu: A mobile face presentation attack database with
  real-world variations,''
\newblock in {\em IEEE international conference on automatic face \& gesture
  recognition}. IEEE, 2017.

\bibitem{zhang2012face}
Zhiwei Zhang et~al.,
\newblock ``A face antispoofing database with diverse attacks,''
\newblock in {\em 2012 5th IAPR international conference on Biometrics (ICB)}.
  IEEE, 2012.

\bibitem{zhang2020celeba}
Yuanhan Zhang et~al.,
\newblock ``Celeba-spoof: Large-scale face anti-spoofing dataset with rich
  annotations,''
\newblock in {\em ECCV}. Springer, 2020.

\bibitem{he2016deep}
Kaiming He et~al.,
\newblock ``Deep residual learning for image recognition,''
\newblock in {\em CVPR}, 2016, pp. 770--778.

\bibitem{selvaraju2017grad}
Ramprasaath~R Selvaraju et~al.,
\newblock ``Grad-cam: Visual explanations from deep networks via gradient-based
  localization,''
\newblock in {\em ICCV}, 2017, pp. 618--626.

\end{thebibliography}
\end{document}


\sloppy

\def\x{{\mathbf x}}
\def\L{{\cal L}}

\title{Learning Facial Liveness Representation \\ for Domain Generalized Face Anti-Spoofing \\
Supplementary Material}
%
\name{Anonymous ICME submission}
\address{}

\maketitle
\renewcommand{\thesection}{\Alph{section}}
\renewcommand{\thefigure}{\Alph{figure}}
\renewcommand{\thetable}{\Alph{table}}
\begin{algorithm}[ht]
\caption{Learning of our framework for domain-generalized facial liveness representation and FAS}
\begin{algorithmic}
\REQUIRE
\STATE \textbf{Input:}
face images $X$, binary labels $Y$, liveness encoder $E_L$, content encoder $E_C$, domain encoder $E_D$, liveness classifier $C_L$, content decoder $D_C$, domain classifier $C_D$
\WHILE{not done} 
\STATE Calculate objectives from Eq. (1) to Eq. (6)
\STATE \textbf{Disentangle Facial Content Representation}
\STATE \quad Update $E_{C}$ with $L_{cont}+L^{cnf}_{cont}$\\
\STATE \quad Update $D_{C}$ with $L_{cont}$\\

\STATE \textbf{Disentangle Domain Representation}
\STATE \quad Update $E_{D}$ with $L_{dom}+L^{cnf}_{dom}$
\STATE \quad Update $C_{D}$ with $L_{dom}$\\

\STATE \textbf{Learning of Generalized Liveness Representation}
\STATE \quad Update $E_{L}$ with $L_{live}+L^{cnf}_{live}$
\STATE \quad Update $C_{L}$ with $L_{live}$\\

\ENDWHILE
\RETURN $E_L, C_L$
\end{algorithmic}
\label{ap:algo}

\end{algorithm}
\section{Implementation Details}
\subsection{Training Algorithm}
\begin{table}[ht]
\begin{center}
\setlength{\tabcolsep}{0.1mm}{
\begin{tabular}{l|l|l|c||clllll}
\hline
\multicolumn{4}{c||}{\textbf{Content Decoder}}                                                  & \multicolumn{6}{c}{\textbf{Liveness Classifier}}                                               \\ \hline
\multicolumn{1}{c|}{Layer} & \multicolumn{2}{c|}{Chan./Stri.} & \multicolumn{1}{c||}{Outp.Size} & \multicolumn{2}{c|}{Layer}    & \multicolumn{2}{c|}{Intp.Size} & \multicolumn{2}{c}{Outp.Size}  \\ \hline
conv-trs-1                 & \multicolumn{2}{c|}{512/2}       & 2                              & \multicolumn{2}{c|}{linear}   & \multicolumn{2}{c|}{1000}      & \multicolumn{2}{c}{512}        \\
batchnorm                  & \multicolumn{2}{c|}{-/-}         & 2                              & \multicolumn{2}{c|}{relu}     & \multicolumn{2}{c|}{512}       & \multicolumn{2}{c}{512}        \\
relu                       & \multicolumn{2}{c|}{-/-}         & 2                              & \multicolumn{2}{c|}{linear}   & \multicolumn{2}{c|}{512}       & \multicolumn{2}{c}{2}          \\ \cline{5-10} 
conv-trs-2                 & \multicolumn{2}{c|}{512/2}       & 4                              & \multicolumn{1}{c}{}    &     &                &               &                &               \\ \cline{5-10} 
batchnorm                  & \multicolumn{2}{c|}{-/-}         & 4                              & \multicolumn{6}{c}{\textbf{Domain Classifier}}                                                 \\ \cline{5-10} 
relu                       & \multicolumn{2}{c|}{-/-}         & 4                              & \multicolumn{2}{c|}{Layer}    & \multicolumn{2}{c|}{Intp.Size} & \multicolumn{2}{c}{Outp.Size} \\ \cline{5-10} 
conv-trs-3                 & \multicolumn{2}{c|}{256/2}       & 8                              & \multicolumn{2}{c|}{linear}   & \multicolumn{2}{c|}{1000}      & \multicolumn{2}{c}{512}       \\
batchnorm                  & \multicolumn{2}{c|}{-/-}         & 8                              & \multicolumn{2}{c|}{relu}     & \multicolumn{2}{c|}{512}       & \multicolumn{2}{c}{512}       \\
relu                       & \multicolumn{2}{c|}{-/-}         & 8                              & \multicolumn{2}{c|}{linear}   & \multicolumn{2}{c|}{512}       & \multicolumn{2}{c}{S}         \\ \cline{5-10} 
conv-trs-4                 & \multicolumn{2}{c|}{128/2}       & 16                             & \multicolumn{1}{c}{}    &     &                &               &                &               \\
batchnorm                  & \multicolumn{2}{c|}{-/-}         & 16                             & \multicolumn{1}{c}{}    &     &                &               &                &               \\
relu                       & \multicolumn{2}{c|}{-/-}         & 16                             & \multicolumn{1}{c}{}    &     &                &               &                &               \\
conv-trs-5                 & \multicolumn{2}{c|}{64/2}        & 32                             & \multicolumn{1}{c}{}    &     &                &               &                &               \\
batchnorm                  & \multicolumn{2}{c|}{-/-}         & 32                             & \multicolumn{1}{c}{}    &     &                &               &                &               \\
relu                       & \multicolumn{2}{c|}{-/-}         & 32                             & \multicolumn{1}{c}{}    &     &                &               &                &               \\
conv-trs-6                 & \multicolumn{2}{c|}{1/2}         & 64                             & \multicolumn{1}{c}{}    &     &                &               &                &               \\ \cline{1-10}
\end{tabular}}
\end{center}
\caption{The detailed structure of the content decoder, the liveness classifier, and the domain classifier.}
\label{ap:stru}
\end{table}
With the introduced objectives discussed in Section 2, we now detail the training process of our proposed network architecture. Given training image samples from $S$ source domains, we train our face content encoder $E_C$ and decoder $D_C$ with $L_{cont}$ for deriving liveness-irrelevant content features, while $L^{cnf}_{cont}$ is calculated from the outputs of the liveness classifier $C_L$ and domain classifier $C_D$ for preserving the content-only information in the derived feature. As for the domain encoder $E_D$ and classifier $C_D$, we calculate $L_{dom}$ and additionally observe $L^{cnf}_{dom}$ to ensure the learning of liveness-irrelevant domain features. Finally, the liveness encoder $E_L$ and classifier $C_L$ are trained by observing $L_{live}$ and $L^{cnf}_{live}$, while the latter is utilized to ensure the disentangling of domain information from the resulting liveness representation. The aforementioned network modules are jointly trained (in an end-to-end learning fashion), and the pseudocode for the training process is summarized in Algorithm~\ref{ap:algo}.\\

Once the training process of our network architecture is complete, only $E_L$ and $C_L$ are utilized for performing domain-generalized FAS. That is, the liveness encoder $E_L$ is applied to extract the domain-invariant liveness representation from the input image, which is fed into $C_L$ for FAS prediction.

\subsection{Training Details}
Our experiment was implemented in PyTorch. We normalize all the face images to $256\times256\times3$ pixels as the input of our proposed model. The output sizes of both the decoder $D_C$, and the PRNet are $64 \times 64 \times 1$ pixels. We set the batch size to 10 and utilized the AdamW optimizer \cite{loshchilov2017decoupled} with a learning rate of 1.5e-4 for training. The proposed model is trained with 200 epochs. We use ResNet-18 \cite{he2016deep} pre-trained on ImageNet \cite{5206848} for the encoders $E_L$, $E_C$ and $E_D$, where the weights of the first layer are shared. The architectures of the liveness classifier $C_L$, content decoder $D_C$, and domain classifier $C_D$ are detailed in Table~\ref{ap:stru}.


\begin{table}[ht]
\begin{center}
\setlength{\tabcolsep}{0.1mm}{
\begin{tabular}{c|c|c|c|c}
    \hline
    \textbf{}&
    \makecell{\textbf{}}&
    \multicolumn{2}{|c|}{\makecell{\textbf{No. of files.}}}&
    \makecell{\textbf{}}\cr
    \cline{3-4}
    \textbf{Dataset}&
    \makecell{\textbf{PA types}}&
    Train&
    Test&
    \makecell{\textbf{Display devices}}\cr
    \hline
        \makecell{OULU~\cite{boulkenafet2017oulu}\\(denoted as O)}& {\makecell{Printed photo \\Display photo\\ Replayed video}}& 1800 & 1800 &\makecell{Dell 1905FP\\ Macbook Retina}\\
    \hline
        \makecell{CASIA~\cite{zhang2012face}\\(denoted as C)}& \makecell{Printed photo \\Cut photo\\ Replayed video}& 240& 360& iPad\\
    \hline
        \makecell{MSU~\cite{wen2015face}\\ (denoted as M)}& \makecell{Printed photo\\ Replayed video}& 120& 160& \makecell{iPad Air\\ iPhone 5S}\\
    \hline
        \makecell{Idiap~\cite{chingovska2012effectiveness}\\ (denoted as I)}& \makecell{Printed photo\\ Display photo\\ Replayed video}& 360& 480& \makecell{iPhone 3GS\\ iPad}\\
    \hline
        \makecell{CelebA-Spoof~\cite{zhang2020celeba}\\ (denoted as Cb)}& \makecell{Printed photo\\ Cut photo\\ Replayed video\\ 3D mask}& $\times$& 20358& \makecell{PC\\ Tablet\\ Phone}\\
    \hline
\end{tabular}}
\end{center}
\caption{Data statistics of the five experimental datasets.}
\label{ap:data}
\end{table}
For the liveness classifier $C_L$ and domain classifier $C_D$, since the extracted liveness feature is expected to be representative for liveness information of an image, we simply apply two linear layers with one ReLU activation layer for our classifier. As for the content decoder $D_C$, we use transpose convolution layers to decode the facial content feature, so as to make sure the content encoder $E_C$ learns the desired feature.

\begin{table}[t]
\begin{center}
\setlength{\tabcolsep}{1mm}{
\begin{tabular}{c|c|c|c|c}
\hline
                                  & \multicolumn{2}{c|}{{\color[HTML]{333333} \textbf{M\&I to C}}} & \multicolumn{2}{c}{\textbf{M\&I to O}} \\ \cline{2-5} 
\multirow{-2}{*}{\textbf{Method}} & HTER(\%)                        & AUC(\%)                         & HTER(\%)             & AUC(\%)             \\ \hline\hline
MS\_LBP~\cite{maatta2011face}                           & 51.16                           & 52.09                           & 43.63                & 58.07               \\
IDA~\cite{wen2015face}                               & 45.16                           & 58.80                           & 54.52                & 42.17               \\
Color Texture~\cite{boulkenafet2016face}                     & 55.17                           & 47.89                           & 53.31                & 45.16               \\
LBPTOP~\cite{de2014face}                            & 45.27                           & 54.88                           & 47.26                & 50.21               \\
MADDG~\cite{shao2019multi}                             & 41.02                           & 64.33                           & 39.35                & 65.10               \\
SSDG-M~\cite{jia2020single}                            & 31.89                           & 71.29                           & 36.01                & 66.88               \\
Cross~\cite{wang2020cross}                             & 31.67                           & 75.23                           & 34.02                & 72.65               \\
DRDG~\cite{liu2021dual}                              & 31.28                           & 71.50                           & 33.35                & 69.14               \\ \hline\hline
\textbf{Ours}                     & \textbf{25.09}                  & \textbf{80.15}                  & \textbf{31.99}       & \textbf{77.14}      \\ \hline
\end{tabular}}
\end{center}
\vspace{-0.2cm}
\caption{Comparisons of FAS in unseen domains using limited source domain data. Following \cite{shao2019multi,jia2020single,wang2020cross,liu2021dual}, images from two source domains as selected for training, while a target domain unseen during training is applied for evaluation. 
}
\label{tab:mm}
\end{table}
\subsection{Experimental Datasets}
Five public face anti-spoofing datasets are used to evaluate the effectiveness of our method: OULU-NPU \cite{boulkenafet2017oulu} (denoted as O), CASIA-FASD \cite{zhang2012face} (denoted as C), Idiap Replay-Attack \cite{chingovska2012effectiveness} (denoted as I), MSU-MFSD \cite{wen2015face} (denoted as M), and CelebA-Spoof \cite{zhang2020celeba} (denoted as Cb). The detailed statistics of each dataset are shown in Table~\ref{ap:data}. The data in O, C, M, and I mainly contains three types of presentation attacks: Printed photo, display photo, and replayed video. As for Cb, other than the presentation attacks mentioned above, it also contains an unseen attack, which is the 3D mask. This dataset is used to verify the performance of our proposed model when seeing an unseen attack from an unseen dataset. 

\section{Extended Experiments}
To further verify the domain generalization capability of our method, we evaluate our model when only two source domains can be observed during training. Following \cite{shao2019multi,jia2020single,wang2020cross,liu2021dual}, we have M\&I as the source domains and select one dataset from C\&O as the target domain. When trained under limited source domains, it is more challenging to perform domain generalized FAS since the model is more likely to over-fit to source datasets. As shown in Table~\ref{tab:mm}, our approach outperforms all the existing FAS approaches by a wide margin in the two tasks. It is expected since our model puts more emphasis on the liveness-relevant feature of the face images with the assist of the facial content disentanglement. With the facial content such as the head poses various in each domain, the existing FAS approaches are not able to perform efficient training when the training data is insufficient. On the other hand, with the facial content disentangled, our model is able to extract generalized liveness features and thus achieves impressive results.

\bibliography{icme}
\bibliographystyle{IEEEbib}